\documentclass[letterpaper]{article} 
\usepackage{aaai25}  
\usepackage{times}  
\usepackage{helvet}  
\usepackage{courier}  
\usepackage[hyphens]{url}  
\usepackage{graphicx} 
\usepackage{natbib}  
\usepackage{caption} 
\usepackage{algorithm}
\usepackage{algorithmic}

\usepackage{newfloat}
\usepackage{listings}
\DeclareCaptionStyle{ruled}{labelfont=normalfont,labelsep=colon,strut=off} 
\floatstyle{ruled}
\newfloat{listing}{tb}{lst}{}
\floatname{listing}{Listing}
\title{Few-shot Metric Domain Adaptation: Practical Learning Strategies for an Automated Plant Disease Diagnosis}
\author {
    Shoma Kudo\textsuperscript{\rm 1},
    Satoshi Kagiwada\textsuperscript{\rm 2},
    Hitoshi Iyatomi\textsuperscript{\rm 1}
}
\affiliations {
    \textsuperscript{\rm 1}Applied Informatics, Graduate School of Science and Engineering, Hosei University, Tokyo, Japan\\
    \textsuperscript{\rm 2}Clinical Plant Science, Faculty of Bioscience and Applied Chemistry, Hosei University, Tokyo, Japan\\
    shoma.kudo.3n@stu.hosei.ac.jp, kagiwada@hosei.ac.jp, iyatomi@hosei.ac.jp
}

\usepackage{bibentry}

\usepackage{multirow}
\usepackage{arydshln}
\usepackage{amsmath}

\begin{document}

\maketitle

\begin{abstract}
    Numerous studies have explored image-based automated systems for plant disease diagnosis, demonstrating impressive diagnostic capabilities. However, recent large-scale analyses have revealed a critical limitation: that the diagnostic capability suffers significantly when validated on images captured in environments (domains) differing from those used during training. This shortfall stems from the inherently limited dataset size and the diverse manifestation of disease symptoms, combined with substantial variations in cultivation environments and imaging conditions, such as equipment and composition. These factors lead to insufficient variety in training data, ultimately constraining the system's robustness and generalization.
To address these challenges, we propose Few-shot Metric Domain Adaptation (FMDA), a flexible and effective approach for enhancing diagnostic accuracy in practical systems, even when only limited target data is available. FMDA reduces domain discrepancies by introducing a constraint to the diagnostic model that minimizes the "distance" between feature spaces of source (training) data and target data with limited samples.
FMDA is computationally efficient, requiring only basic feature distance calculations and backpropagation, and can be seamlessly integrated into any machine learning (ML) pipeline. In large-scale experiments, involving 223,015 leaf images across 20 fields and 3 crop species, FMDA achieved F1 score improvements of 11.1 to 29.3 points compared to cases without target data, using only 10 images per disease from the target domain. Moreover, FMDA consistently outperformed fine-tuning methods utilizing the same data, with an average improvement of 8.5 points.
\end{abstract}

\begin{figure*}[t]
    \centering
    \includegraphics[width=0.8\linewidth]{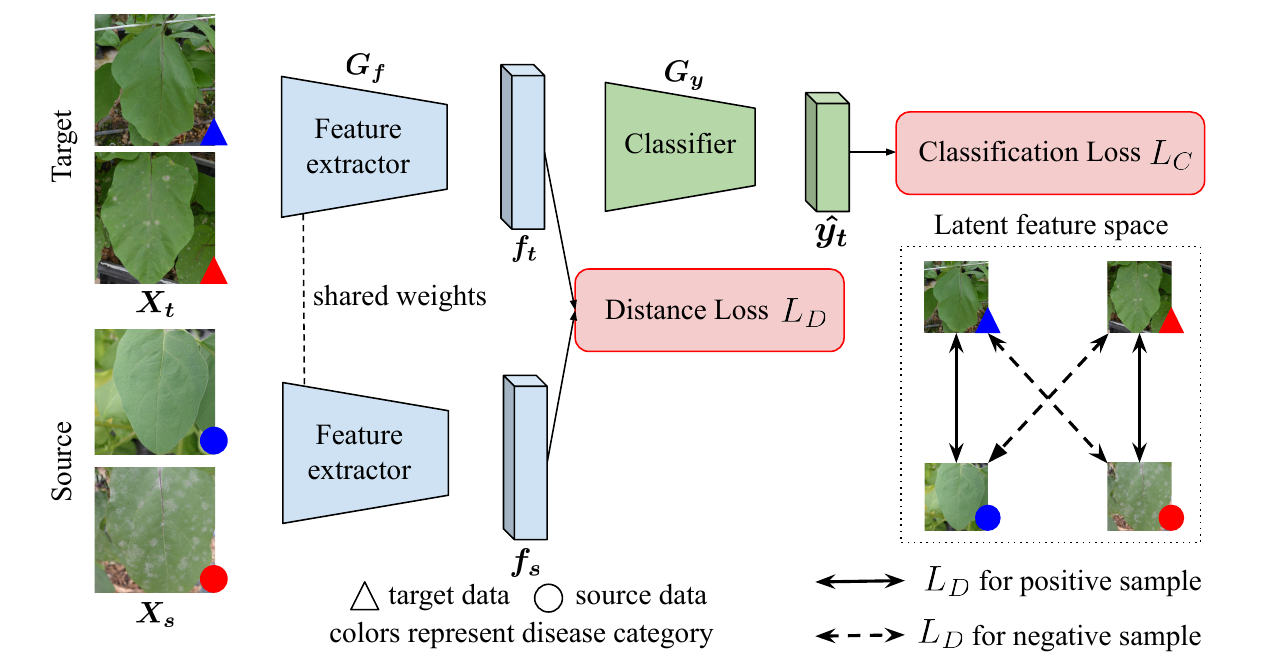}
    \caption{\textbf{Architecture of FMDA.} Triangles ($\triangle$) represent target domain data, circles ($\bigcirc$) represent source domain data, and colors indicate the data labels (disease classes).}
    \label{fig:FMDA}
\end{figure*}

\section{Introduction}
    Plant diseases cause significant damage to crops, posing a major challenge for farmers worldwide. According to the Food and Agriculture Organization (FAO), 20–40\% of global food crops are lost annually to plant pests and diseases, resulting in economic losses estimated at approximately \$290 billion per year \cite{food2019new}. Early detection of diseases in plants is critical to minimizing these losses. However, current diagnostic methods rely heavily on subjective visual inspections by experts, which may not always be accurate, especially in cases of subtle or complex disease symptoms. This highlights the pressing need for a simple yet highly accurate automated plant disease diagnosis system.

Deep learning techniques, particularly convolutional neural networks (CNNs), have demonstrated remarkable performances in plant disease diagnosis, achieving high classification accuracy in numerous studies \cite{mohanty2016using, hiroki2018diagnosis, ferentinos2018deep, atila2021plant}. However, subsequent large-scale and rigorous evaluations have revealed a significant drop in diagnostic performance when these models are validated on unseen datasets, captured under conditions different from those used during training \cite{mohanty2016using, boulent2019convolutional, shibuya2021validation}. In many earlier studies, exceptionally high accuracy was reported due to inappropriate train-test data splitting, since images within the same domain (environment) shared similar characteristics. When training and test data were probabilistically split without accounting for domain diversity, implicit data leakage occurred.
Shibuya et al. addressed this issue by conducting a large-scale analysis using a dataset of 221,000 images of four crops collected from 24 prefectures, ensuring careful management of shooting fields to prevent data leakage. While they achieved accuracy levels comparable to prior studies (averaging 99\%) on data from the same fields used for training, accuracy on data from different fields ranged from 64\% to 88\% \cite{shibuya2021validation}. This discrepancy is attributed to the inherently small and diverse nature of disease symptoms, coupled with significant variations in visual appearance, cultivation methods, and photography conditions across fields (domains) and datasets. The limited diversity of training data, even with tens of thousands of images, fails to capture these variations, hindering the generalization of diagnostic performance to unseen domains. As such, the fundamental diagnostic capability of a system should be evaluated on data unseen to the model, from unknown domains. However, even state-of-the-art machine learning (ML) techniques struggle to achieve high accuracy on such scenarios, disadvantaged by limited diversity in training data.

Domain adaptation techniques have shown promising results for tasks where training and target data exhibit domain differences \cite{ganin2016domain, tzeng2017adversarial, dinsdale2021deep, li2023domain}. These methods are particularly valuable for large-scale tasks involving data integration from multiple domain sources. Although domain adaptation has been applied to plant disease diagnosis with some success \cite{yan2021deep,fuentes2021open,wu2023laboratory}, existing approaches—often based on unsupervised domain adaptation (UDA)—require substantial amounts of unlabeled target data. Furthermore, their effectiveness diminishes when the distribution gap between source (training) and target (test) domains is too large \cite{wang2019characterizing,zhao2019learning}.

Efforts to address the limited diversity of training data include augmenting datasets using generative models. Techniques leveraging Generative Adversarial Networks (GANs) have achieved promising results by generating high-quality images \cite{nazki2020unsupervised,cap2020leafgan,kanno2021ppig}. However, since GAN-generated images are derived from existing training data, they face limitations in scenarios with significant domain discrepancies. Recent advances in diffusion models \cite{ho2020denoising}, particularly latent diffusion models \cite{rombach2022high}, enable more flexible and diverse image generation by leveraging extensive pre-trained image datasets and textual information. These advancements hold potential for augmenting training datasets in plant disease diagnosis tasks.

Given the challenge of accurately diagnosing diseases in unseen domains with limited training data diversity, this study considers a practical scenario where a small amount of data from the target diagnostic domain is available. We aim to effectively utilize this limited data to improve diagnostic accuracy.

In this paper, we propose Few-shot Metric Domain Adaptation (FMDA), a supervised domain adaptation method that leverages a small amount of target data effectively. FMDA introduces a learning constraint to reduce the "distance" between feature spaces of the target and source domains, thereby enhancing the model's ability to generalize. By extending conventional fine-tuning techniques with FMDA, we enable stable mitigation of domain shifts while utilizing existing knowledge from the source domain. FMDA is a simple and versatile approach applicable to various tasks with significant domain shifts. It is particularly effective for tasks like plant disease diagnosis, where intra-class variability is extremely high. In this paper, we demonstrate the application of FMDA to automated plant disease diagnosis, assuming scenarios where limited data from the target imaging environment is available. We validate its effectiveness through large-scale experiments.

The contributions of this study are as follows:
\begin{enumerate}
\item We propose FMDA, a supervised domain adaptation method that can be applied to various tasks with serious domain shifts using limited target data.
\item We develop a highly accurate automated plant disease diagnosis system capable of diagnosing target data with high precision.
\end{enumerate}

\section{Related Works}
    \subsection{Improving Model Generalizability in Automated Plant Disease Diagnosis}

\subsubsection{Region of Interest (ROI) detection}
Saikawa et al. proposed a system that removes background features, which are considered significant sources of variability across fields, reporting an approximate accuracy improvement of 12\% \cite{saikawa2019aop}. However, subsequent experiments by Shibuya et al., using a larger dataset with higher-resolution images, revealed that the impact of background removal on diagnostic performance was limited \cite{shibuya2021validation}.
In contrast, methods utilizing object detection techniques have been reported to be more robust to variations in subject distance while providing higher interpretability of results \cite{suwa2019comparable, chen2022plant}. Specifically, Wayama et al. demonstrated through large-scale analysis that ROI detection is particularly effective for diagnosing pest infestations, which are smaller in size but exhibit less variability compared to plant diseases \cite{wayama2024investigation}.

\subsubsection{Data Generation}
Nazki et al. proposed AR-GAN, an extension of CycleGAN \cite{zhu2017unpaired} that incorporates a loss function to maintain critical feature similarity between real and generated images. Experiments using a tomato leaf dataset showed that the inclusion of generated images during training enhanced the model's classification accuracy \cite{nazki2020unsupervised}.
Quan et al. introduced LeafGAN, which ensures that disease symptoms are localized specifically within leaf regions, rather than appearing in the background, enabling the generation of high-quality images \cite{cap2020leafgan}.
Kanno et al. proposed Productive and Pathogenic Image Generation (PPIG), a two-stage generative process that first produces healthy leaf images and then overlays disease symptoms. This approach generates more diverse images, with disease symptoms being the only variable feature within otherwise identical compositions \cite{kanno2021ppig}.
While these GAN-based methods have contributed to improving classification accuracy, the generated images are inherently constrained by the distribution of the original training data, reducing their effectiveness in addressing fundamental domain diversity issues. In contrast, latent diffusion models \cite{rombach2022high} can generate more diverse images compared to GANs. Although still relatively new, these models hold great promise for advancing data augmentation techniques in plant disease diagnosis \cite{muhammad2023harnessing}.

\subsection{Studies on Domain Adaptation in ML Tasks}

\subsubsection{General Unsupervised Domain Adaptation (UDA)}
Domain adaptation aims to reduce the distributional gap (domain shift) between the "source domain" (training data) and the "target domain" (test data), thereby enhancing model performance on the target domain.
In many real-world scenarios, test data cannot be used during training. A practical alternative is the transductive learning setup, where unlabeled test data is fully available for training, but the model is not expected to generalize beyond this scope. This setup has driven extensive research into unsupervised domain adaptation (UDA).
Ganin et al. proposed the Domain Adversarial Neural Network (DANN), which simultaneously trains a domain discriminator and a primary classification task using source data. By reversing the gradient of the domain discriminator during backpropagation, DANN enforces domain-invariant feature extraction \cite{ganin2016domain}.
Tzeng et al. introduced Adversarial Discriminative Domain Adaptation (ADDA), which adversarially trains a target domain encoder while feeding its outputs into a domain discriminator. This approach constructs a domain-invariant encoder, leveraging a pre-trained source domain encoder \cite{tzeng2017adversarial}. These adversarial approaches have been widely adopted across various ML application.

\subsubsection{UDA in Automated Plant Disease Diagnosis}
In the context of automated plant disease diagnosis, UDA has been applied in a limited number of studies.
Fuentes et al. proposed a diagnostic framework that applies UDA to the bounding boxes of detected disease symptoms, mitigating domain shifts in a dataset of tomato leaf images captured across three fields \cite{fuentes2021open}.
Wu et al. applied UDA to a training framework designed to learn diverse feature representations using a dataset comprising leaf images of various crops captured in both laboratory and outdoor environments \cite{wu2023laboratory}.
While these methods have demonstrated potential for improving diagnostic performance, they rely heavily on the availability of large amounts of unlabeled target data. Furthermore, when the data distributions of the source and target domains differ significantly, these methods often fail to bridge the domain gap effectively, thereby limiting diagnostic performance.

To address these challenges, this study considers a practical scenario where a small amount of labeled target data is available. We propose Few-shot Metric Domain Adaptation (FMDA), a domain adaptation method that effectively leverages this limited data to achieve superior diagnostic performance.
        
\section{FMDA: Few-shot Metric Domain Adaptation}
    \subsection{Problem Setting and Overview of FMDA}

In this study, the labeled training dataset consists of $C$ classes of source data $\{X_s, y_s\}$ and a small number of labeled target data $\{X_t, y_t\}$ for the same $C$ classes. Here, $X$ represents the input data (e.g., images), and $y$ represents the corresponding class labels. While the source data is available in sufficient quantities, the target data is limited to $n$ samples per class, amounting to only $C \times n$ samples in total.
The primary objective is to effectively leverage this limited labeled target data to train an ML model $G$ capable of achieving high diagnostic accuracy on unseen target data.

We propose Few-shot Metric Domain Adaptation (FMDA), a diagnostic framework designed to address challenging tasks, such as automated plant disease diagnosis, where domain discrepancies are significant. FMDA is both practical and capable of achieving excellent diagnostic performance.
The architecture of FMDA is illustrated in Figure \ref{fig:FMDA}. The ML model $G$ in FMDA consists of a feature extractor $G_f$ and a classifier $G_y$, and the choice of models for $G_f$ and $G_y$ is flexible.
FMDA training comprises two phases:
\begin{enumerate}
\item \textbf{Pre-training Phase:} The model $G$ is pre-trained on the source data $\{X_s, y_s\}$ using supervised learning.
\item \textbf{Adaptation Phase:} The pre-trained model $G$ is fine-tuned using the limited $C \times n$ target domain training samples $\{X_t, y_t\}$. This phase integrates metric learning into a standard fine-tuning strategy, enabling more effective utilization of the scarce but valuable target data to construct a robust model.

Conventional fine-tuning applied in scenarios with extremely limited target data risks degrading the original model $G$ and causing overfitting. On the other hand, re-training the entire model using both source and target data is computationally expensive. Moreover, due to the significantly smaller number of target samples compared to source data, the overall impact of such additional training is often limited.
FMDA addresses these challenges through a simple yet effective combination of fine-tuning and metric learning. This approach ensures that the model not only retains the general knowledge learned from the source domain but also adapts effectively to the unique characteristics of the target domain, even with minimal data. \end{enumerate}

\subsection{Implementation Details of FMDA}

Since the pre-training phase performs general training of the model using the source data $\{X_s, y_s\}$ as mentioned above, the adaptation phase of the FMDA is explained here.
In the adaptation phase, the entire loss function $L$ the model to be minimized is defined as:
\begin{equation}
    L = L_C(X_t,y_t) + \lambda L_D .
    \label{eq:FMDA_loss}
\end{equation}
Here, $L_C$ is a classification loss based on the commonly used cross entropy, computed only by target data $\{X_t,y_t\}$:
\begin{equation}
    L_C(X_t, y_t) = - \sum^N_{i=1} y_t^{(i)} \log(G_y(G_f(X_t^{(i)}))) .
    \label{eq:classification_loss}
\end{equation}
Here, $N$ denotes the batch size used during training.

The term $L_D$ corresponds to the distance-based loss, which is introduced to mitigate overfitting and domain shift. The hyperparameter $\lambda$ balances the contribution of classification loss and distance-based loss.
Conventional fine-tuning often suffers from overfitting when only a small amount of target data is available. This makes the model difficult to generalize to unseen target data. By incorporating $L_D$, FMDA leverages information from both source and target domains, ensuring that the feature representations of target data are properly aligned with the source feature space. This alignment reduces domain discrepancies while preserving robust classification boundaries.

In this study, we define two types of $L_D$:
\begin{enumerate}
\item \textbf{Euclidean Distance Loss} ($L_{D,\text{L2}}$): This loss minimizes the Euclidean distance between the feature representations of target samples and positive samples ($X_{s_p}^{(i)}$: same labels from $X_t$).
\begin{equation}
\hspace{-5mm} 
L_{D,\text{L2}} = \dfrac{1}{N} \sum_{i=1}^N \big\| G_f(X_t^{(i)}) - G_f(X_{s_p}^{(i)}) \big\|_2 .
\label{eq:l2_loss}
\end{equation}
\item \textbf{Improved Triplet Loss} ($L_{D,\text{Triplet+}}$): This loss enhances feature separation by maximizing the distance between target samples and negative samples ($X_{s_n}^{(i)}$: different labels from $X_t$) while minimizing the distance to positive samples.
\begin{equation}
\hspace{-5mm} 
\begin{split}
L_{D,\text{Triplet+}} = & \dfrac{1}{N} \sum_{i=1}^N \biggl\{
\Bigl[ \bigl\|G_f(X_t^{(i)}) - G_f(X_{s_p}^{(i)})\bigr\|_2 \\
&- \bigl\|G_f(X_t^{(i)}) - G_f(X_{s_n}^{(i)})\bigr\|_2 + \alpha \Bigr]_+ \\
&+ \Bigl[ \bigl\|G_f(X_t^{(i)}) - G_f(X_{s_p}^{(i)})\bigr\|_2 - \beta \Bigr]_+ \biggr\} .
\end{split}
\label{eq:triplet+_loss}
\end{equation}

$\alpha$ and $\beta$ are margin parameters, and $[z]_+ = \text{max}(0, z)$.
\end{enumerate}

The choice of distance function in FMDA has some influence on its performance, depending on the specific conditions. The Euclidean distance loss is computationally efficient and performs well in cases where intra-class variability is low, with features being tightly clustered. However, its performance may decrease in scenarios where intra-class variability is high, leading to greater diversity within the same class. In contrast, The Improved triplet loss offers additional benefits by minimizing the distance between positive pairs (target and source data with the same label) while increasing the separation between negative pairs (target and source data with different labels). This property makes it more effective in cases with high inter-class similarity or overlapping feature distributions.
Practitioners using FMDA should weigh the computational cost against the complexity of feature distributions in the target domain. For applications requiring efficiency, Euclidean Distance Loss may be a practical choice, while Improved Triplet Loss could be advantageous for tasks demanding more discriminative feature spaces.

By incorporating $L_D$ into FMDA, the method effectively enhances generalization by balancing fine-tuning with domain adaptation. This integration enables FMDA to handle domain-shift-heavy tasks with improved flexibility and robustness.

\section{Experiments}
    \subsection{Experimental Setup}

\begin{figure}[t]
    \centering
    \includegraphics[width=1.0\linewidth]{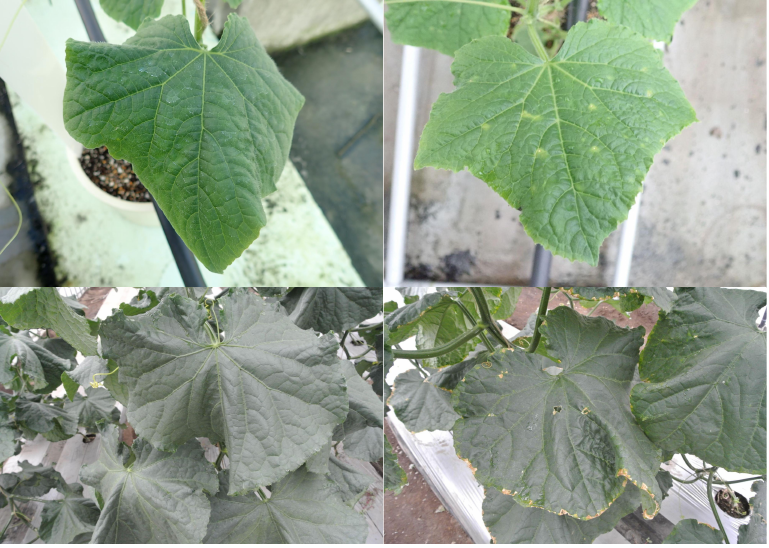}
    \caption{Examples of domain differences in the cucumber dataset. \textsl{HEALTHY\_Source} (top-left), \textsl{Bacterial\_Spot\_Source} (top-right), \textsl{HEALTHY\_Target} (bottom-left), \textsl{Bacterial\_Spot\_Target} (bottom-right).}
    \label{fig:domain_dif}
\end{figure}

To evaluate the effectiveness of FMDA, we benchmarked it against the automated plant disease diagnosis task, a highly relevant application area, and conducted large-scale experiments using our proprietary dataset. The dataset comprises 223,015 images collected from 20 fields, covering 3 crops and a total of 30 disease classes. Details of the dataset are shown in Table \ref{table:dataset}.
The dataset was split into source (Training) and target (Test) data for different farms taken or different periods of time. From the target data, only a very small number of samples ($n$ samples per disease class) were used for training. As shown in Figure \ref{fig:domain_dif}, the domains differ significantly in features such as cultivation environments, backgrounds, and compositions.

In this study, EfficientNetV2-S \cite{tan2021efficientnetv2} was employed as the feature extractor $G_f$, and the classifier $G_y$ was implemented as a simple fully connected layer with a softmax activation function throughout all phases of experiments.
In the \textbf{pre-training phase}, the model was initialized with weights pre-trained on ImageNet1K \cite{russakovsky2015imagenet} dataset. Subsequently, pre-training was conducted using the source domain data $\{X_s,y_s\}$. The following conditions were applied during this phase:
\begin{itemize}
  \item \textbf{Optimizer:} Adam \cite{kingma2014adam} with a learning rate of 0.001.
  \item \textbf{Batch Size:} 128.
  \item \textbf{Data Augmentation:} Applied transformations included random cropping to $480 \times 480$ pixels, random 90-degree rotations, horizontal flips, and brightness adjustments.
\end{itemize}
In the \textbf{adaptation phase}, the model weights initialized during the pre-training phase were further fine-tuned using the target data $\{X_t,y_t\}$. The experimental setup for this phase is as follows:
\begin{itemize}
  \item \textbf{Optimizer:} Adam with a learning rate of 0.001.
  \item \textbf{Batch Size:} $2 \times C$, where $C$ represents the number of classes in the target data.
  \item \textbf{Data Augmentation:} Identical to the pre-training phase.
\end{itemize}
The primary focus was on evaluating the impact of domain adaptation learning. Therefore, only basic data augmentation were applied. Based on preliminary experiments, the hyperparameter $\lambda$, controlling the balance between the classification loss and the distance-based loss, was set to $\lambda = 1$.

\begin{table}[t]
\caption{Dataset details.}
\label{table:dataset}
\centering
\scalebox{0.9}{
\begin{tabular}{llrr}
\hline
Crop                       & ID\_Name                      & Source & Target \\ \hline
\multirow{12}{*}{Cucumber} & 00\_HEALTHY                   & 16,016 & 5,583  \\
                           & 01\_Powdery\_Mildew           & 7,757  & 1,905  \\
                           & 02\_Gray\_Mold                & 636    & 174    \\
                           & 03\_Anthracnose               & 3,031  & 80     \\
                           & 08\_Downy\_Mildew             & 6,946  & 2,586  \\
                           & 09\_Corynespora\_Leaf\_Spot   & 7,557  & 1,820  \\
                           & 17\_Gummy\_Stem\_Blight       & 1,476  & 381    \\
                           & 20\_Bacterial\_Spot           & 4,355  & 2,655  \\
                           & 22\_CCYV                      & 5,961  & 186    \\
                           & 23\_Mosaic\_diseases          & 26,854 & 1,633  \\
                           & 24\_MYSV                      & 17,229 & 1,011  \\ \cline{2-4} 
                           & Total                         & 97,818 & 18,014 \\ \hline
\multirow{8}{*}{Eggplant}  & 00\_HEALTHY                   & 12,431 & 1,122  \\
                           & 01\_Powdery\_Mildew           & 7,936  & 938    \\
                           & 02\_Gray\_Mold                & 1,024  & 166    \\
                           & 06\_Leaf\_Mold                & 3,188  & 732    \\
                           & 11\_Leaf\_Spot                & 5,509  & 119    \\
                           & 18\_Verticillium\_Wilt        & 3,176  & 354    \\
                           & 19\_Bacterial\_Wilt           & 3,415  & 463    \\ \cline{2-4} 
                           & Total                         & 36,679 & 3,894  \\ \hline
\multirow{13}{*}{Tomato}   & 00\_HEALTHY                   & 8,120  & 2,994  \\
                           & 01\_Powdery\_Mildew           & 4,490  & 4,250  \\
                           & 02\_Gray\_Mold                & 9,327  & 571    \\
                           & 05\_Cercospora\_Leaf\_Mold    & 4,078  & 1,809  \\
                           & 06\_Leaf\_Mold                & 2,761  & 151    \\
                           & 07\_Late\_Blight              & 2,049  & 808    \\
                           & 10\_Corynespora\_Target\_Spot & 1,732  & 1,350  \\
                           & 19\_Bacterial\_Wilt           & 2,259  & 412    \\
                           & 21\_Bacterial\_Canker         & 4,353  & 128    \\
                           & 27\_ToMV                      & 3,453  & 49     \\
                           & 28\_ToCV                      & 4,320  & 871    \\
                           & 29\_Yellow\_Leaf\_Curl        & 4,513  & 1,746  \\ \cline{2-4} 
                           & Total                         & 51,471 & 15,139 \\ \hline
\end{tabular}
}
\end{table}

\subsection{Evaluation}

We evaluated the diagnostic performance of the classifier $G$ on test data (11 cucumber classes, 7 eggplant classes, and 12 tomato classes) using the F1 score. The following methods were compared:

\begin{enumerate}
\item \textbf{w/o Target data:} No information from target data is used.
\item \textbf{DANN:} The DANN framework \cite{ganin2016domain} is applied in a transductive setting where all target domain images are available as unlabeled data during training. 
\item \textbf{finetune:} $n$ samples per disease from target data are used for fine-tuning.
\item \textbf{DANN-tune:} The representative UDA method DANN \cite{ganin2016domain} is adapted to fine-tuning with $n$ samples per disease. 
\item \textbf{FMDA(L2):} The proposed method using Euclidean distance for $L_D$. 
\item \textbf{FMDA(Triplet+):} The proposed method using Improved Triplet Loss \cite{cheng2016person} for $L_D$.
\end{enumerate}

For these experiments, the number of target data samples was set to $n=3$ and $n=10$ per disease in each crop. Each experiment was repeated five times with different target data, and results were evaluated diagnosis performance.
The number of training iterations for the pre-training and adaptation phases was set to 100 and 1,000, respectively, based on preliminary experiments. In addition, since the diagnostic performance of some methods varies with the number of training iterations, the performance at the point of maximum test performance was reported along with the corresponding number of iterations.
To qualitatively evaluate the extent to which each method reduces domain differences, the feature distribution $f$ of both domain data obtained from the feature extractor $G_f$ was visualized using t-SNE \cite{van2008visualizing}.

\section{Results}
    \begin{table*}[t]
\centering 
\caption{Comparison of plant disease diagnosis performance at the 1,000th epoch in macro F1 (\%: mean $\pm$ SD).}
\label{table:result_1000}
\begin{tabular}{lllllll}
\hline
Crop                   & \multicolumn{2}{c}{Cucumber} & \multicolumn{2}{c}{Eggplant}     & \multicolumn{2}{c}{Tomato} \\ \hline
\# added samples ($n$ / class) &
  \multicolumn{1}{c}{3} &
  \multicolumn{1}{c}{10} &
  \multicolumn{1}{c}{3} &
  \multicolumn{1}{c}{10} &
  \multicolumn{1}{c}{3} &
  \multicolumn{1}{c}{10} \\ \hline
1. w/o Target data          & \multicolumn{2}{c}{48.0}     & \multicolumn{2}{c}{72.7}         & \multicolumn{2}{c}{48.0}   \\ 
2. DANN          & \multicolumn{2}{c}{47.9}     & \multicolumn{2}{c}{67.9}         & \multicolumn{2}{c}{44.5}   \\ \hdashline
3. finetune (baseline) & 69.73±2.69    & 70.05±2.79   & 78.81±3.56          & 78.39±2.15 & 62.54±5.39   & 73.72±2.42   \\
4. DANN-tune           & 61.45±4.04    & 69.08±1.99   & 73.99±2.75          & 78.89±2.14 & 62.61±3.78   & 67.30±2.91   \\
5. FMDA(L2) (ours)     & 72.11±2.62    & 74.57±1.53   & \textbf{81.61±1.03} & 82.73±3.03 & 65.75±1.53   & 74.19±2.78   \\
6. FMDA(Triplet+) (ours) &
  \textbf{72.91±1.45} &
  \textbf{77.29±1.84} &
  80.62±2.90 &
  \textbf{83.83±1.97} &
  \textbf{65.91±2.76} &
  \textbf{74.37±1.42} \\ \hline
\end{tabular}
\\ \raggedright{Scores in bold indicate best results in the category.}
\end{table*}
\begin{table*}[t]
\centering 
\caption{Reference results: Comparison of plant disease diagnosis performance at the best-performing epoch (epoch count) in macro F1 (\%: mean $\pm$ SD).}
\label{table:result_best}
\scalebox{0.83}{
\hspace{-5mm} 
\tabcolsep = 4pt
\begin{tabular}{lrrrrrrrrrrrr}
\hline
Crop                   & \multicolumn{4}{c}{Cucumber}            & \multicolumn{4}{c}{Eggplant}                    & \multicolumn{4}{c}{Tomato}              \\ \hline
\multicolumn{1}{r}{\# added samples ($n$ / class)} &
  \multicolumn{2}{c}{3} &
  \multicolumn{2}{c}{10} &
  \multicolumn{2}{c}{3} &
  \multicolumn{2}{c}{10} &
  \multicolumn{2}{c}{3} &
  \multicolumn{2}{c}{10} \\ \hline
3. finetune (baseline) & 69.85±1.42 & (200) & 76.32±1.50 & (120) & 81.04±1.44          & (600) & 87.17±1.70 & (50) & 65.24±2.40 & (400) & 74.85±2.41 & (180) \\
4. DANN-tune           & 68.82±1.67 & (100) & 75.64±0.81 & (20)  & 77.71±3.12          & (500) & 86.53±1.17 & (10) & 62.86±3.75 & (900) & 74.85±2.41 & (180) \\
5. FMDA(L2) (ours)     & 72.38±1.73 & (700) & 79.39±1.57 & (400) & \textbf{85.36±1.39} & (400) & 89.08±1.63 & (50) & 70.74±2.02 & (200) & 79.26±0.75 & (90)  \\
6. FMDA(Triplet+) (ours) &
  \textbf{73.79±2.30} &
  (400) &
  \textbf{79.90±1.31} &
  (300) &
  85.20±2.04 &
  (500) &
  \textbf{89.15±1.11} &
  (30) &
  \textbf{70.79±2.58} &
  (200) &
  \textbf{80.38±1.51} &
  (60) \\ \hline
\end{tabular}
}
\\ \raggedright{Scores in bold indicate best results in the category.}
\end{table*}
\begin{figure}[t]
    \hspace{-7mm} 
    \includegraphics[width=1.07\linewidth]{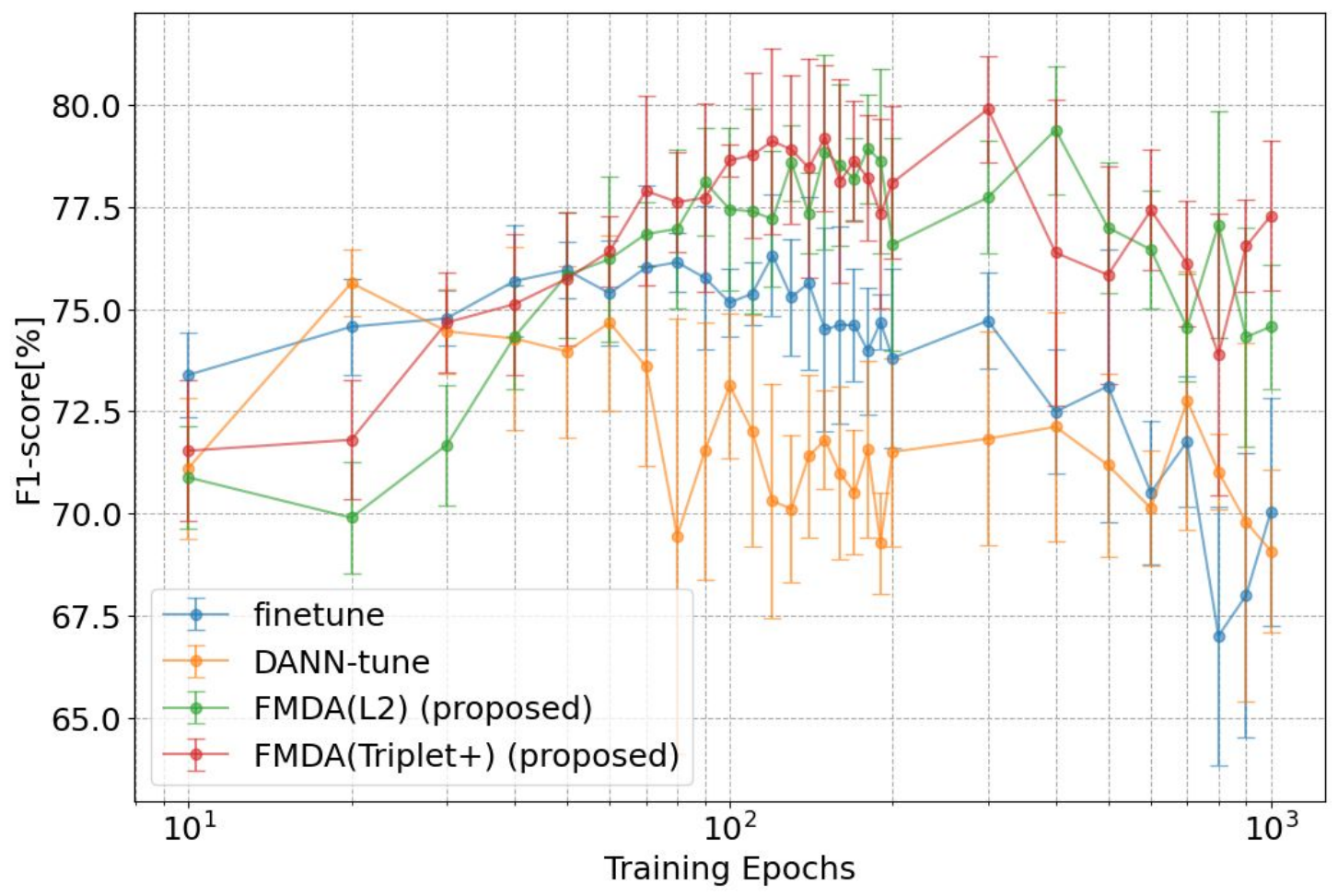}
    \caption{Performance trends for the cucumber dataset.}
    \label{fig:scores_cucumber}
\end{figure}

Table \ref{table:result_1000} summarizes the diagnostic performance of the models after 1,000 training iterations. Although the dependence of the diagnostic performance on the number of training iterations varied by method, the observed trends were generally consistent across all crops. As a representative example, Figure \ref{fig:scores_cucumber} shows the relationship between the number of training iterations and the diagnosis performance for the cucumber dataset when $n=10$.

Table \ref{table:result_best} the F1 scores achieved at the best-performing number of training iterations for each method. This provides additional insight into the peak performance potential of each approach.

Figure \ref{fig:embedding_tomato} visualizes the feature distribution of the tomato dataset’s source and target data using t-SNE, at $n=10$. The visualization highlights how well each method aligns the target domain features with the source domain, offering a qualitative measure of domain adaptation effectiveness.
    
\section{Discussion}
    \begin{figure*}[t]
    \centering
    \includegraphics[width=1.0\linewidth]{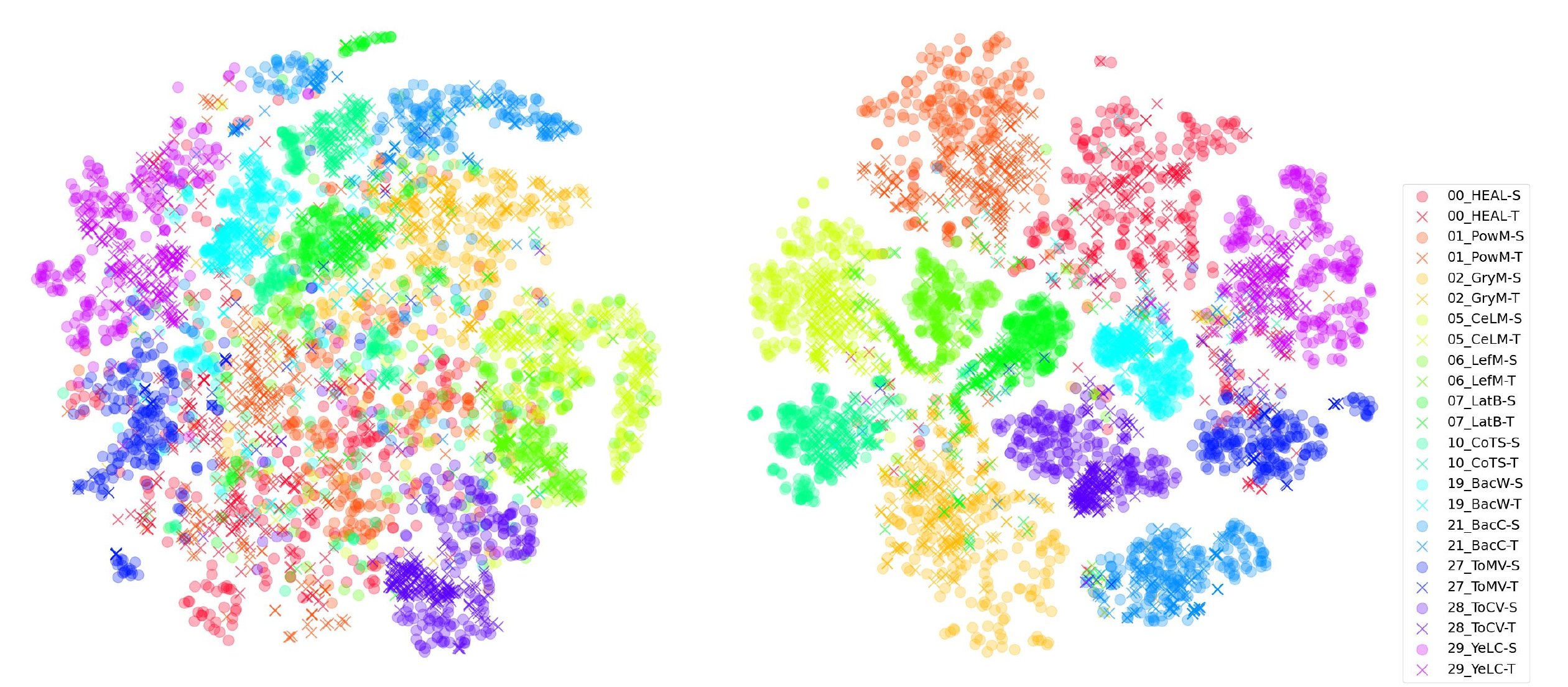}
    \caption{Feature distribution for the tomato dataset. DANN-tune (left), FMDA(Triplet+) (right). Circles ($\bigcirc$) represent source data, and crosses ($\times$) represent target data.}
    \label{fig:embedding_tomato}
\end{figure*}

When target data is not utilized, the diagnostic performance remains extremely low, even when state-of-the-art ML classifiers are trained on 30,000–90,000 images per crop. This reaffirms that automated plant disease diagnosis is inherently a domain-shift-heavy problem, making it challenging to achieve practical diagnostic accuracy without addressing domain discrepancies.

Among the methods leveraging label information from the test data, DANN-tune exhibited the lowest diagnostic performance. DANN reduces domain discrepancy by training the feature extractor $G_f$ to produce domain-invariant features. However, this approach does not guarantee that target data with the same labels as source data will be mapped to nearby locations in the feature space. As illustrated in the feature distribution (Figure \ref{fig:embedding_tomato}), DANN-tune incorrectly maps target data into inappropriate class distributions, particularly in datasets with significant domain shifts.
Furthermore, the DANN (Transductive Learning) setting, which utilized all unlabeled target domain images, also performed worse than the baseline w/o Target. This counterintuitive result suggests that using unlabeled target data without proper alignment of features can exacerbate domain shifts, leading to degraded diagnostic performance.
These findings highlight that UDA approaches, such as DANN, are ineffective for tasks with substantial domain discrepancies, such as the current task of automated plant disease diagnosis.

In contrast, FMDA consistently achieved high diagnostic performance regardless of the distance loss function used. This was evident not only in its practical performance after a fixed number of training iterations but also in its best-performing potential, where FMDA demonstrated a higher upper limit of achievable performance compared to existing methods. Figure \ref{fig:embedding_tomato} demonstrates that FMDA effectively leverages target label information to map target features appropriately onto the source feature space. This indicates that FMDA can successfully learn from limited target data, reducing domain discrepancies while preserving diagnostic accuracy.

Incorporating as few as three labeled target data samples per disease resulted in an average F1 score improvement of 9.8–16.9 points across methods, while using ten target samples yielded an average improvement of 15.5–22.3 points. Although performance improves as more target data becomes available, even minimal amounts — such as three samples per disease — significantly enhance diagnostic performance. This underscores the importance of utilizing even limited labeled target data in domain-shift-heavy tasks like automated plant disease diagnosis.

Fine-tuning with a small amount of target data experienced a performance drop of 6.3 points due to overfitting, whereas FMDA exhibited only a modest decline of approximately 2.5 points. This suggests that fine-tuning is highly susceptible to overfitting in low-data scenarios, while FMDA effectively mitigates overfitting and facilitates stable learning.

While FMDA demonstrates significant improvements in diagnostic performance, certain limitations remain. For example, its reliance on labeled target data, albeit in small quantities, may not be feasible in real-world scenarios where labeling is costly or impractical.
Future research could investigate integrating FMDA with semi-supervised learning techniques to harness a larger pool of unlabeled data. Additionally, while this study focuses on plant disease diagnosis, FMDA's framework could be extended to other domain-shift-heavy tasks, such as medical imaging or remote sensing. Validating FMDA in these domains would further assess its generalizability and robustness under varied conditions.
    
\section{Conclusion}
    In this study, we proposed Few-shot Metric Domain Adaptation (FMDA), a supervised domain adaptation method designed to address tasks with significant domain shifts. FMDA demonstrated its ability to effectively and reliably learn from limited target domain information, even in scenarios where UDA-based approaches fail due to large domain discrepancies.
    
\section{Acknowledgement}
    This work was supported by the Ministry of Agriculture, Forestry and Fisheries (MAFF), Japan, under the commissioned project study “Development of Pest Diagnosis Technology Using AI” (JP17935051), and by the Cabinet Office through the Public/Private R\&D Investment Strategic Expansion Program (PRISM).

%

\bibliography{aaai25}

\end{document}